\documentclass{article}
\PassOptionsToPackage{numbers, compress}{natbib}

\usepackage[final]{nips_2017}

\usepackage[utf8]{inputenc} 
\usepackage[T1]{fontenc}    
\usepackage{hyperref}       
\usepackage{url}            
\usepackage{booktabs}       
\usepackage{amsfonts}       
\usepackage{nicefrac}       
\usepackage{microtype}      
\usepackage{graphicx}
\usepackage{subfigure}
\graphicspath{{figures/}}
\usepackage{natbib}

\usepackage{xcolor}

\usepackage{algorithm}
\usepackage{algpseudocode}
\usepackage{bm}
\usepackage{amsmath}
\usepackage{amsfonts}
\usepackage{amssymb}

\input{Definitions}

\ifx\proof\undefined
\newenvironment{proof}{\par\noindent{\bf Proof\ }}{\hfill\BlackBox\\[2mm]}
\fi

\ifx\theorem\undefined
\newtheorem{theorem}{Theorem}
\fi
\ifx\example\undefined
\newtheorem{example}{Example}
\fi
\ifx\lemma\undefined
\newtheorem{lemma}[theorem]{Lemma}
\fi

\ifx\corollary\undefined
\newtheorem{corollary}[theorem]{Corollary}
\fi

\ifx\assumption\undefined
\newtheorem{assumption}{Assumption}
\fi

\ifx\definition\undefined
\newtheorem{definition}{Definition}
\fi

\ifx\proposition\undefined
\newtheorem{proposition}[theorem]{Proposition}
\fi

\ifx\remark\undefined
\newtheorem{remark}{Remark}
\fi

\ifx\conjecture\undefined
\newtheorem{conjecture}[theorem]{Conjecture}
\fi

\ifx\factoid\undefined
\newtheorem{factoid}[theorem]{Fact}
\fi

\ifx\axiom\undefined
\newtheorem{axiom}[theorem]{Axiom}
\fi

\newcommand{\RN}[1]{%
	\textup{\lowercase\expandafter{\it \romannumeral#1}}%
}

\def\KL{\textsf{KL}} 
\def\JSD{\textsf{JSD}} 

\usepackage{wrapfig}
\usepackage{color}
\title{Particle Optimization in Stochastic Gradient MCMC}

\author{
	Changyou Chen \\
	Dept. of Computer Science and Engineering\\
	SUNY at Buffalo\\
	Buffalo, NY 14226 \\
	\texttt{cchangyou@gmail.com} \\
   \And
   Ruiyi Zhang \\
   Dept. of Computer Science \\
   Duke University \\
   Durham, NC 27708 \\
   \texttt{zhangry868@gmail.com} \\
}

\begin{document}

\maketitle

\begin{abstract}
Stochastic gradient Markov chain Monte Carlo (SG-MCMC) has been increasingly popular in Bayesian learning due to its ability to deal with large data. A standard SG-MCMC algorithm simulates samples from a discretized-time Markov chain to approximate a target distribution. However, the samples are typically highly correlated due to the sequential generation process, an undesired property in SG-MCMC. In contrary, Stein variational gradient descent (SVGD) directly optimizes a set of particles, and it is able to approximate a target distribution with much fewer samples. In this paper, we propose a novel method to directly optimize particles (or samples) in SG-MCMC from scratch. Specifically, we propose efficient methods to solve the corresponding Fokker-Planck equation on the space of probability distributions, whose solution ({\it i.e.}, a distribution) is approximated by particles. Through our framework, we are able to show connections of SG-MCMC to SVGD, as well as the seemly unrelated generative-adversarial-net framework. Under certain relaxations, particle optimization in SG-MCMC can be interpreted as an extension of standard SVGD with momentum.
\end{abstract}

\section{Introduction}

Bayesian methods have been playing an important role in modern machine learning, especially in an unsupervised-learning setting. When facing with big data, two lines of research directions have been developed to scale up Bayesian methods, {\it e.g.}, variational-Bayes-based and sampling-based methods. Stochastic gradient Markov chain Monte Carlo (SG-MCMC) is a family of scalable Bayesian learning algorithms designed to efficiently sample from a target distribution such as a posterior distribution \cite{WellingT:ICML11,ChenFG:ICML14,DingFBCSN:NIPS14,ChenDC:NIPS15}. In principle, SG-MCMC generates samples from a Markov chain, which are used to approximate a target distribution. Under a standard setting, samples from SG-MCMC are able to match a target distribution exactly in an infinite-sample regime \cite{TehTV:arxiv14,ChenDC:NIPS15}. However, this case never occurs in practice, as only a finite amount of samples are available. Although nonasymptotic bounds w.r.t.\! the number of samples have been investigated \cite{TehTV:arxiv14,VollmerZT:arxiv15,ChenDC:NIPS15}, there are no theory/algorithms to guide learning an optimal set of fixed-size samples/particles. This is an undesirable property of SG-MCMC, because given a fixed number of samples, one often wants  ot learn the optimal samples that best approximate a target distribution.

A remedy for this issue is to adopt the idea of particle-based sampling methods, where a set of particles (or samples) are initialized from some simple distribution, and they are updated iteratively such that they approximate a target distribution better and better. The updating procedure is usually done by optimizing some objective function. There is not much work in this direction for Bayesian sampling, with an outstanding representative being the Stein variational gradient descent (SVGD) \cite{LiuW:NIPS16}. In SVGD, the update of particles are done by optimizing the KL-divergence between the empirical-particle distribution and a target distribution, thus the samples can be guaranteed to be optimal in each update. Because of this property, SVGD is found to perform better than SG-MCMC when the number of samples used to approximate a target distribution is limited \cite{LiuW:NIPS16}.

Little research has been done on investigating the particle-optimization idea in SG-MCMC.
Inspired by SVGD, we develop a similar particle-optimization procedure for SG-MCMC for more efficient sampling. To achieve this goal, we propose a novel technique to directly optimize particles based on a variational reformulation of the corresponding Fokker-Planck equation of an SG-MCMC algorithm, adapted from \cite{JordanKO:MA98}. In this way, instead of sampling from a Markov chain sequentially, we evolve particles through an optimization procedure, obtaining both optimal particles and faster convergence speed compared to standard SG-MCMC. Furthermore, under some relaxations, we are able to show particle optimization in SG-MCMC can be regarded as an extension of SVGD with momentum. To the best of our knowledge, this is the first time particles can be optimized in SG-MCMC algorithms.

\section{Preliminaries}

\subsection{Stochastic gradient MCMC}
\paragraph{Diffusion-based sampling methods}
Generating random samples from a posterior distribution is a pervasive problem in Bayesian statistics which has many important applications in machine learning. The Markov Chain Monte Carlo method (MCMC), proposed by Metropolis {\em et al.} \cite{Metropolis:53}, produces unbiased samples from a desired distribution when the density function is known up to a normalizing constant. However, traditional MCMC methods are based on random walk proposals which often lead to highly correlated samples. On the other hand, dynamics-based sampling methods, {\it e.g.}, Hybrid Monte Carlo (HMC) \cite{Duane:87,Horowitz:91}, avoid this high degree of correlation by combining dynamical systems with the Metropolis step. The dynamical system uses information from the gradient of the log density to reduce the random walk effect, and the Metropolis step serves as a correction of the discretization error introduced by the numerical integration of the dynamical systems. In fact, these dynamical systems are derived from a more general mathematical technique called diffusion process (or more specifically, It\'{o} diffusion) \cite{Oksendal:85}.

Specifically, our objective is to generate random samples from a posterior distribution $p(\thetab|\Xb)\propto p(\Xb|\thetab)p(\thetab)$, where $\thetab\in\mathbb{R}^r$ represents the model parameter, and $\Xb\triangleq \{\xb_i\}_{i=1}^N$ represents the data. The canonical form is $p(\thetab|\Xb)=(1/Z)\exp(-U(\thetab))$, where
\begin{align}\label{eq:potential}
	U(\thetab)=-\log p(\Xb|\thetab)-\log p(\thetab) \triangleq -\sum_{i=1}^N\log p(\xb_i | \thetab) - \log p(\thetab)
\end{align}
is referred to as the potential energy based on an i.i.d.\! assumption of the model, and $Z$ is the normalizing constant. 
In general, the posterior distribution can be corresponding to the (marginal) stationary distribution of a (continuous-time) It\'{o} diffusion, defined as a stochastic differential equation of the form:
\begin{align}\label{eq:itodif}
	\mathrm{d}\Thetab_t = F(\Thetab_t)\mathrm{d}t + g(\Thetab_t)\mathrm{d}\mathcal{W}_t~,
\end{align}
where $t$ is the time index; $\Thetab_t \in \mathbb{R}^{\pr}$ represents the full variables in a dynamical system, and
$\Thetab_t \supseteq \thetab_t$ (thus $\pr \geq r$) is potentially an augmentation of model parameter $\thetab$; $\mathcal{W}_t \in \mathbb{R}^{\pr}$ is $\pr$-dimensional Brownian motion. Functions $F: \mathbb{R}^{\pr} \to \mathbb{R}^{\pr}$ and $g: \mathbb{R}^{\pr} \rightarrow \mathbb{R}^{\pr\times \pr}$ are assumed to satisfy the Lipschitz continuity condition \cite{Ghosh:book11}. By Fokker-Planck equation (or the forward Kolmogorov equation) \cite{Kolmogoroff:MA31,Risken:FPE89}, when appropriately designing the diffusion-coefficient functions $F(\cdot)$ and $g(\cdot)$, the stationary distribution of the corresponding It\'{o} diffusion equals the posterior distribution of interest, $p(\thetab|\Xb)$. For example, the 1st-order Langevin dynamic defines $\Thetab = \thetab$, and $F(\Thetab_t) = -\nabla_{\thetab} U(\thetab),~~ g(\Thetab_t) = \sqrt{2}\Ib_r$; the 2nd-order Langevin diffusion defines $\Thetab = (\thetab, \qb)$, and 
$F(\Thetab_t)= \Big( \begin{array}{c}
\qb \\
-B \qb-\nabla_\thetab U(\thetab) \end{array} \Big),\hspace{0.1cm}
g(\Thetab_t) = \sqrt{2B}\Big( \begin{array}{cc}
{\bf 0} & {\bf 0} \\
{\bf 0} & \Ib_n \end{array} \Big)$
for a scalar $B > 0$; $\qb$ is an auxiliary variable known as the momentum \cite{ChenFG:ICML14,DingFBCSN:NIPS14}.

Denoting the distribution of $\Thetab_t$ as $\rho_t$, it is well known \cite{Risken:FPE89} that $\rho_t$ is characterized by the Fokker-Planck (FP) equation:
\begin{align}\label{eq:FPE}
	\frac{\partial \rho_t}{\partial t} = -\nabla_{\Thetab}\cdot \left(\rho_tF(\Thetab_t)\right) + \nabla_{\Thetab}\!\!\nabla_{\Thetab}\!:\!\left(\rho_tg(\Thetab_t)g^{\top}(\Thetab_t)\right)~,
\end{align}
where $\ab\cdot\bb \triangleq \ab^{\top} \bb$ for vectors $\ab$ and $\bb$, $\Ab\!:\!\Bb\triangleq \mbox{trace}(\Ab^{\top}\Bb)$ for matrices $\Ab$ and $\Bb$. The FP equation is the key to develop our particle-optimization framework in SG-MCMC.

\paragraph{Stochastic gradient MCMC}
SG-MCMC algorithms are discretized numerical approximations of the It\'{o} diffusions. They mitigate the slow mixing and non-scalability issues encountered by traditional MCMC algorithms by $\RN{1})$ adopting gradient information of the posterior distribution, and $\RN{2})$ using minibatches of the data in each iteration of the algorithm to generate samples. To make the algorithms scalable in a big-data setting, three developments will be implemented based on the It\'{o} diffusion: $\RN{1})$ define appropriate functions $F$ and $g$ in the It\'{o}-diffusion formula so that the (marginal) stationary distributions coincide with the target posterior distribution $p(\thetab|\Xb)$; $\RN{2})$ replace $F$ or $g$ with unbiased stochastic approximations to reduce the computational complexity, {\it e.g.}, approximating $F$ with a random subset of the data instead of using the full data. For example, in the 1st-order Langevin dynamics, $\nabla_{\thetab}U(\thetab)$ could be {\em approximated} by an unbiased estimator with a subset of data:
\begin{align}\label{eq:U}
	\nabla_\thetab \tilde{U}(\thetab) \triangleq \nabla\log p(\thetab)+ \frac{N}{n}\sum_{i=1}^n \nabla_{\thetab}\log p(\xb_{\pi_i}|\thetab)
\end{align}
where $\pi$ is a size-$n$ random subset of $\{1, 2, \cdots, N\}$, leading to the first SG-MCMC algorithm in machine learning -- stochastic gradient Langevin dynamics (SGLD) \cite{WellingT:ICML11}; and $\RN{3})$ solve the generally intractable continuous-time It\^{o} diffusions with a numerical method, {\it e.g.}, the Euler method \cite{ChenDC:NIPS15}. For example, this leads to the following update in SGLD:
\begin{align*}
	\thetab^{(\ell)} = \thetab^{(\ell-1)} - \nabla_{\thetab}\tilde{U}(\thetab^{(\ell-1)})h + \sqrt{2h}\deltab_{\ell}~,
\end{align*} 
where $h$ means the stepsize, $\ell$ indexes the samples, $\deltab_{\ell}\sim\mathcal{N}(\mathbf{0}, \Ib)$ is a random sample from an isotropic normal distribution. After running the algorithm for $L$ steps, the collection of samples $\{\thetab^{(\ell)}\}_{\ell=1}^L$, which are collected from a Markov chain, are used to approximate the unknown posterior distribution $\frac{1}{Z}e^{-U(\thetab)}$.

\subsection{Stein variational gradient descent}
Different from SG-MCMC, SVGD initializes a set of particles and iteratively updates them so that the empirical particle distribution approximates the posterior distribution. Specifically, considers a set of particles $\{\thetab_i\}_{i=1}^M$ drawn from distribution $q$. SVGD tries to update these particles by doing gradient descent on the space of probability distributions via
\begin{equation}\label{eq:svgd}
\begin{aligned}
\thetab_i \leftarrow \thetab_i + \epsilon \phi(\thetab_i),~~\phi = \arg\max_{\phi\in \mathcal{F}} \left\{\dfrac{\partial}{\partial \epsilon} \KL(q_{[\epsilon\phi]}||p)\right\},
\end{aligned}
\end{equation}
where $\phi$ is a function perturbation direction chosen to minimize the KL divergence between the updated empirical distribution $q_{[\epsilon\phi]}$ and the posterior $p(\thetab|\Xb)$, $p$ for short. Since $\KL(q\|p)$ is convex in $q$, global optimum of $q = p$ can be guaranteed.
SVGD considers $\mathcal{F}$ as the unit ball of a vector-valued reproducing kernel Hilbert space (RKHS) $\mathcal{H}$ associated with a kernel $\kappa(\thetab,\thetab^{\prime})$. In such as setting, it was shown in \citep{liu2016stein} that:
\begin{align}\label{eq:ksd}
	-\frac{\partial}{\partial \epsilon} \KL(q_{[\epsilon \phi]}\|p)|_{\epsilon=0} &= \mathbb{E}_{\thetab \sim q}[\text{trace}(\Gamma_p \phi(\thetab))],\\
	\text{with}
	~~\Gamma_p \phi(\thetab) &\triangleq \nabla_{\thetab} \log p( \thetab | \Xb) ^{\top} \phi(\thetab) + \nabla_{\thetab} \cdot \phi(\thetab), \nonumber
	\end{align}
where $\Gamma_p$ is called the Stein operator. 
Assuming that the update function $\phi(\thetab)$ is in a RKHS with kernel $\kappa(\cdot,\cdot)$, it was shown in \citep{liu2016stein} that (\ref{eq:ksd}) is maximized with:
\begin{align}
	\label{equ:close}
	\phi(\thetab) = \mathbb{E}_{\thetab\sim q}[\kappa(\thetab, \thetab^\prime) \nabla_{\thetab} \log p( \thetab |  \Xb)
	+ \nabla_{\thetab} \kappa(\thetab, \thetab^\prime)].
\end{align}
When approximating the expectation $\mathbb{E}_{\thetab\sim q}[\cdot]$ with empirical particle distribution, we arrive at the following updates for the particles at the $\ell$-th iteration:
\begin{align}  \label{eq:svgd_update}
\begin{aligned}
\thetab_i^{(\ell)} = \thetab_i^{(\ell-1)} + \dfrac{\epsilon}{M} \sum_{j=1}^M \left[ \kappa(\thetab_j^{(\ell-1)}, \thetab_i^{(\ell-1)}) \nabla_{\thetab_j} \log p( \thetab_j^{(\ell-1)} |  \Xb)
+ \nabla_{\thetab_j} \kappa(\thetab_j^{(\ell-1)}, \thetab_i^{(\ell-1)}) \right].
\end{aligned}
\end{align}
SVGD applies the updates in (\ref{eq:svgd_update}) repeatedly, and the samples move closer to the target distribution $p$ in each iteration.

\subsection{Comparing SG-MCMC with SVGD}
SG-MCMC is a Markov-chain-based sampling methods, in the sense that samples are generated from a Markov chain, with potentially highly correlated samples. Furthermore, it often requires a large number of samples in order to approximate a target distribution reasonably well \cite{ChenDC:NIPS15}. In contrast, SVGD directly updates the particles to their optimum guided by an objective function, thus requires much less samples to approximate a target distribution.

On the other hand, SVGD has been explained as gradient flows whose gradient operator is defined on the RKHS \cite{liu2017stein_flow}; whereas SG-MCMC are flows with flow operator defined on the $\mathcal{L}_2$ space - the functional space that is square integrable. Since  RKHS is smaller than $\mathcal{L}_2$, SG-MCMC can potentially obtain better asymptotic properties than SVGD in theory \cite{liu2017stein_flow}.

The above arguments motivate us to combine goods from both sides, {\it i.e.}, we aim to developed a particle-based SG-MCMC algorithm similar to what SVGD does.

\section{Particle Optimization in SG-MCMC}

To develop our particle-optimization framework, we first introduce the following lemma adapted from \cite{ChenLCWPC:arxiv17}, viewing SG-MCMC from an optimization perspective.

\begin{lemma}\label{lem:variational_fp}
	Assume that $p(\thetab_t|\Xb)\leq C_1$ is infinitely differentiable, and $\|\nabla_{\thetab}\log p(\thetab|\Xb)\| \leq C_2\left(1 + C_1 - \log p(\thetab|\Xb)\right) (\forall \thetab)$ for some constants $\{C_1, C_2\}$. Let $T = \epsilon K$ with $K$ being an integer, $\tilde{\rho}_0$ is an arbitrary distribution with same support as $p(\thetab|\Xb)$, and $\{\tilde{\rho}_k\}_{k=1}^K$ be the solution of the functional optimization problem:
	\begin{align}\label{eq:variationalFP}
		\tilde{\rho}_k = \arg\min_{\rho \in \mathcal{K}}\mbox{KL}\left(\rho \| p(\thetab|\Xb)\right) + \frac{1}{2h}W^2_2\left(\tilde{\rho}_{k-1}, \rho\right)~,
	\end{align}
	where $W_2^2\left(\mu_1, \mu_2\right) \triangleq \inf_{p\in \mathcal{P}(\mu_1, \mu_2)}\int \left\|\xb - \yb\right\|_2^2p(\mathrm{d}\xb,\mathrm{d}\yb)$, $W_2\left(\mu_1, \mu_2\right) $ is the 2nd-order Wasserstein distance, with $\mathcal{P}(\mu_1, \mu_2)$ being the space of joint distributions of $\{\mu_1, \mu_2\}$; $\mathcal{K}$ is the space of probability distributions with finite 2nd-order moment. Then $\tilde{\rho}_K$ converges to $\rho_T$ in the limit of $h\rightarrow 0$, {\it i.e.}, $\lim_{h\rightarrow 0}\tilde{\rho}_K = \rho_T$, where $\rho_T$ is the solution of the FP equation \eqref{eq:FPE} at time $T$.
\end{lemma}

Lemma~\ref{lem:variational_fp} reveals an interesting way to compute $\rho_T$ via a sequence of functional optimization problems. By comparing it with the objective of SVGD, which minimizes the KL-divergence between $\rho_k$ and $p(\thetab | \Xb)$, at each sub-optimization-problem in Lemma~\ref{lem:variational_fp}, it minimizes the KL-divergence, plus a regularization term as the Wasserstein distance between $\tilde{\rho}_{k-1}$ and $\tilde{\rho}_k$. The extra Wasserstein-distance term arises naturally due to the fact that the corresponding diffusion is a gradient flow equipped with a geometric associated with the Wasserstein distance \citep{Otto:ARMA98}. From another point of view, it is known that the Wasserstein distance is a better metric for probability distributions than the KL-divergence, especially in the case of non-overlapping domains \citep{ArjovskyB:ICLR17,ArjovskyCB:arxiv17}.

According to Lemma~\ref{lem:variational_fp}, it is now apparent that SG-MCMC can be achieved by alternatively solving the optimization problem in \eqref{eq:variationalFP} for each iteration.

\subsection{Optimizing on the space of probability distributions}
Our idea of particle-optimization is inspired by Lemma~\ref{lem:variational_fp}, in that we can obtain the optimal distribution $\tilde{\rho}_k$ for each iteration (which will be approximated by particles) by optimizing \eqref{eq:variationalFP}. Consequently, instead of doing simulation based on the original It\'{o} diffusion, we propose to directly solve \eqref{eq:variationalFP} on the space of probability distributions $\mathcal{K}$. As will be shown, this allows us to derive algorithms which directly optimize particles in SG-MCMC. However, this also bring challenges for the optimization, as the probability-distribution space $\mathcal{K}$ is too flexible to derive exact solutions. In the following, we propose techniques to approximate the corresponding terms in \eqref{eq:variationalFP}. We denote $\thetab_k$ as a sample from $\tilde{\rho}_k$, {\it i.e.}, $\thetab_k \sim \tilde{\rho}_k$.

\subsubsection{Approximating the KL-divergence}
Given $\tilde{\rho}_{k-1}$, the solution $\tilde{\rho}_k$ of \eqref{eq:variationalFP} can be considered as an unknown transformation $\mathcal{G}$ from $\tilde{\rho}_{k-1}$ to $\tilde{\rho}_k$, {\it i.e.},
\begin{align}\label{eq:transform}
	\tilde{\rho}_k = \mathcal{G}(\tilde{\rho}_{k-1})~,
\end{align}
under the constraint that $\tilde{\rho}_k$ still lies in $\mathcal{K}$ after the transformation. Directly solving the unknown transformation is challenging, we propose two methods to solve it approximately, detailed below.

\paragraph{Optimizing $\mathcal{G}$ with adversarial learning}
Optimizing the KL divergence with an unknown transformation $\mathcal{G}$ is generally infeasible. Instead, we approximate it with the Jensen-Shanon divergence, which appears to have the same optimality solution, thus they are equivalent. Formally, we first introduce the following lemma.

\begin{lemma}\label{lem:kl_jsd}
	Let $p_1$ and $p_2$ be probability distributions on $(\mathcal{X}, \Sigma)$ with the same support. Then KL divergence $\KL(p_1\|p_2)$ is equivalent to the Jensen-Shanon divergence (JSD) $\JSD(p_1\|p_2)$ in the sense that they are both convex in $(p_1, p_2)$, and achieve the same minimum value of zero at $p_1 = p_2$. 
\end{lemma}

Based on Lemma~\ref{lem:kl_jsd}, we can replace the KL term in \eqref{eq:variationalFP} with the JSD, resulting in
\begin{align}\label{eq:kl2jsd}
	\tilde{\rho}_k = \arg\min_{\rho \in \mathcal{K}}\mbox{JSD}\left(\rho \| p(\thetab|\Xb)\right) + \frac{\lambda}{2h}W^2_2\left(\tilde{\rho}_{k-1}, \rho\right)~,
\end{align}
where $\lambda > 0$ is introduced to balance the difference between the scales of KL and JSD.

It is well known that JSD is the metric for measuring the distance between two probability distributions in generative adversarial networks (GANs) \cite{GoodfellowAMXWOCB:NIPS14}. According to the properties of GAN, it is easy to see that the unknown transformation $\mathcal{G}$ is equivalent to the generator network in GANs. The only difference is that in our case, the latent space of GAN will be the same as the data space, which does not impact the learning algorithm. Consequently, we can update the transformation $\mathcal{G}$ in each iteration by running a GAN update on its generator, which are then used to generate samples from $\tilde{\rho}_k$.

\paragraph{Optimizing $\mathcal{G}$ via kernelized Stein operator}
Optimizing $\mathcal{G}$ with adversarial learning described above brings us a little computation overhead to update the transformation. Here we propose a more efficient method based on the kernelized Stein operator \cite{LiuW:NIPS16}. We first introduce a theoretical result from \cite{LiuW:NIPS16} in Lemma~\ref{lem:KLdir}.

\begin{lemma}[\cite{LiuW:NIPS16}]\label{lem:KLdir}
	Let $\mathcal{H}$ denotes the reproducing kernel Hilbert space (RKHS), and $\mathcal{H}^r$ the space of vector functions $\fb = [f_1, \cdots, f_r]$ with $f_i \in \mathcal{H}$. Let $\mathcal{G}(\thetab) = \thetab + \fb(\thetab)$, where $\fb\in \mathcal{H}^r$. Denote $\rho$ to be the density of $\tilde{\thetab} = \mathcal{G}(\thetab)$, then we have
	\begin{align*}
		\underset{\fb \in \mathcal{H}^r}{\nabla_{f}\KL(\rho\|p(\thetab|\Xb))}\left|_{\fb = 0}\right. = -\phi^*_{\rho,p}(\thetab)~,
	\end{align*}
	where $\phi^*_{\rho,p}(\thetab) = \mathbb{E}_{\thetab\sim \rho}\left[k(\thetab, \cdot)\nabla_{\thetab}\log p(\thetab|\Xb) + \nabla_{\thetab}k(\thetab, \cdot)\right]$.
\end{lemma}

Lemma~\ref{lem:KLdir} essentially says that the functional gradient of the KL divergence is $ -\phi^*_{q,p}(\thetab)$, when the transformation is in the form of $\mathcal{G}(\thetab) = \thetab + \fb(\thetab)$ with $\fb$ restricted to $\mathcal{H}^r$. The result seems to be applicable in our problem \eqref{eq:variationalFP} except that we require $f_i$ to be in a large space $\mathcal{L}_2$ instead of $\mathcal{H}$. To compromise this, we propose to inject noise into the functional gradient, leading to the following approximation:
\begin{align}\label{eq:steino}
	\underset{\fb \in \mathcal{L}^r}{\nabla_{f}\KL(\rho\|p(\thetab|\Xb))}\left|_{\fb = 0}\right. \approx -\phi^*_{\rho,p}(\thetab) + \sigma \deltab~,
\end{align}
where $\mathcal{L}_2^r \triangleq \mathcal{L}_2\times \cdots\times \mathcal{L}_2$, $\deltab \sim \mathcal{N}(\mathbf{0}, \Ib)$, and $\sigma$ controls the variance of the injected noise, which is typically decreasing in the algorithm to ensure convergence. Note this resembles stochastic gradient descent, where full gradients are replaced with noisy gradients, thus convergence can still be guaranteed.

\subsubsection{Approximating the Wasserstein distance}

In order to calculate the Wasserstein term on the RHS of \eqref{eq:variationalFP}, we adapt results from optimal transport theory \cite{Villani:08,FeiziSXT:arxiv17} to rewrite $W^2_2\left(\tilde{\rho}_{k-1}, \rho\right)$ as
\begin{align}\label{eq:w2}
	&W^2_2\left(\tilde{\rho}_{k-1}, \rho\right) \\
	=& \mathbb{E}_{\tilde{\rho}_{k-1}}\left\|\thetab_{t-1}\right\|^2 + \mathbb{E}_{\rho}\left\|\thetab\right\|^2 + 2\sup_{\psi(\cdot) \text{ convex }}-\mathbb{E}_{\tilde{\rho}_{k-1}}\left[\psi(\thetab_{t-1})\right] - \mathbb{E}_{\rho}\left[\psi^*(\thetab)\right]~,\nonumber
\end{align}
where $\psi^*(\thetab) \triangleq \sup_{\vb}\left(\vb^T\thetab - \psi(\vb)\right)$ is the convex-conjugate of the function $\psi$. 

Optimizing $W^2_2\left(\tilde{\rho}_{k-1}, \rho\right)$ is in general infeasible due to the need to search on the space of convex functions. However, we can approximate it by restricting $\psi$ on some nice convex functions. We describe some preliminary results below.

\paragraph{Restricting $\psi$ to be quadratic}
We can defined $\psi$ to be in the form of 
\begin{align*}
	\psi(\thetab) \triangleq \frac{1}{2}\langle \thetab, \Ab\thetab\rangle + \langle \bb, \thetab\rangle + c
\end{align*}
with parameters $\{\Ab, \bb, c\}$. In this case, the convex-conjugate is in a nice form of 
\begin{align*}
	\psi^*(\thetab) \triangleq \frac{1}{2}\langle \thetab - \bb, \Ab^{-1}(\thetab - \bb)\rangle - c~.
\end{align*}
Substituting these forms into the ``$\sup$'' part and maximizing w.r.t.\! $\{\Ab, \bb, c\}$ by setting the derivatives to zero, we have
\begin{align*}
	\bb = \thetab - \Ab\thetab_{t-1}
\end{align*}
Substituting the above formula into \eqref{eq:w2} and simplifying, we have
\begin{align}\label{eq:w2quad}
	W^2_2\left(\tilde{\rho}_{k-1}, \rho\right)
	&= \mathbb{E}_{\tilde{\rho}_{k-1}}\left\|\thetab_{t-1}\right\|^2 + \mathbb{E}_{\rho}\left\|\thetab\right\|^2 - 2\mathbb{E}_{\tilde{\rho}_{k-1},\rho}\left[\thetab_{t-1}^T\thetab\right] \nonumber\\
	& = \mathbb{E}_{\tilde{\rho}_{k-1},\rho}\left(\thetab - \thetab_{t-1}\right)^2~.
\end{align}

\paragraph{Restricting $\psi$ to other forms}
It is also interesting to parameterize $\psi$ with other convex functions. We are currently investigating on this direction.

\subsection{Particle optimization}
The above sections describes how to optimize the distribution $\tilde{\rho}_k$ directly from \eqref{eq:variationalFP}. In practice, one must adopt some representation for $\tilde{\rho}_k$ in order to perform optimization. A simple representation is to use particles for approximation, {\it i.e.}, $\tilde{\rho}_k \approx \frac{1}{M}\sum_{i=1}^M\delta_{\thetab_i^{(k)}}$ where $\delta_{\thetab}$ is a point mass at $\thetab$, $\{\thetab_i^{(k)}\}$ is the set of particles at the $k$-th iteration. 

Denote our objective to be $\mathcal{F}(\{\thetab_i^{(\ell)}\})$ where $\ell$ indexes the iteration. We can update the particles by calculating their gradients and perform gradient descent. For example, when adopting the approximations \eqref{eq:steino} and \eqref{eq:w2quad}, we have
\begin{align}\label{eq:sgmcmc_grad}
	\frac{\partial}{\partial \thetab_i^{(\ell)}}\mathcal{F}(\{\thetab_i^{(\ell)}\}) \approx -\hat{\phi}^*(\thetab_i^{(\ell)}) + \sigma_{\ell} \deltab + \frac{1}{h}(\thetab_i^{(\ell)} - \thetab_i^{(\ell-1)})~,
\end{align}
where $\hat{\phi}^*(\thetab) = \frac{1}{M}\sum_{i=1}^M\left[k(\thetab_i^{(\ell)}, \thetab)\nabla_{\thetab_i^{(\ell)}}\log p(\thetab_i^{(\ell)}|\Xb) + \nabla_{\thetab_i^{(\ell)}}k(\thetab_i^{(\ell)}, \thetab)\right]$.

\paragraph{SG-MCMC vs SVGD with Polyak's momentum gradient descent}
Let $f(\thetab)$ be the objective function to be optimized. Polyak's momentum gradient descent update \cite{Polyak:CMMP64} is given by
\begin{align}\label{eq:polya}
	\thetab^{(\ell)} = \thetab^{(\ell-1)} - h \nabla_{\thetab}f(\thetab) + \mu\left(\thetab^{(\ell-1)} - \thetab^{(\ell-2)}\right)~.
\end{align}

To see the relation of particle optimization in SG-MCMC of \eqref{eq:sgmcmc_grad} with SVGD, first note that because \eqref{eq:w2} is an upper bound for the 2nd-order Wasserstein distance, we should scale it by some constant $\mu\in(0, 1)$ in the implementation to approximate the true $W^2_2\left(\tilde{\rho}_{k-1}, \rho\right)$. Based on the gradient formula in \eqref{eq:sgmcmc_grad}, the update equation for $\thetab_i^{(\ell)}$ then becomes
\begin{align*}
	\thetab_i^{(\ell)} = \thetab_i^{(\ell-1)} - h\left(\hat{\phi}^*(\thetab_i^{(\ell-1)}) + \sigma_{\ell} \deltab\right) + \mu\left(\thetab_i^{(\ell-1)} - \thetab_i^{(\ell-2)}\right)~,
\end{align*}
which is in the same form of \eqref{eq:polya} where the gradient is represented as noisy gradient. Thus particle optimization in SG-MCMC with \eqref{eq:sgmcmc_grad} can be regarded as SVGD with Polyak's momentum method.

\section{Empirical Verification}

\begin{figure}
	\centering
	\includegraphics[width=0.48\linewidth]{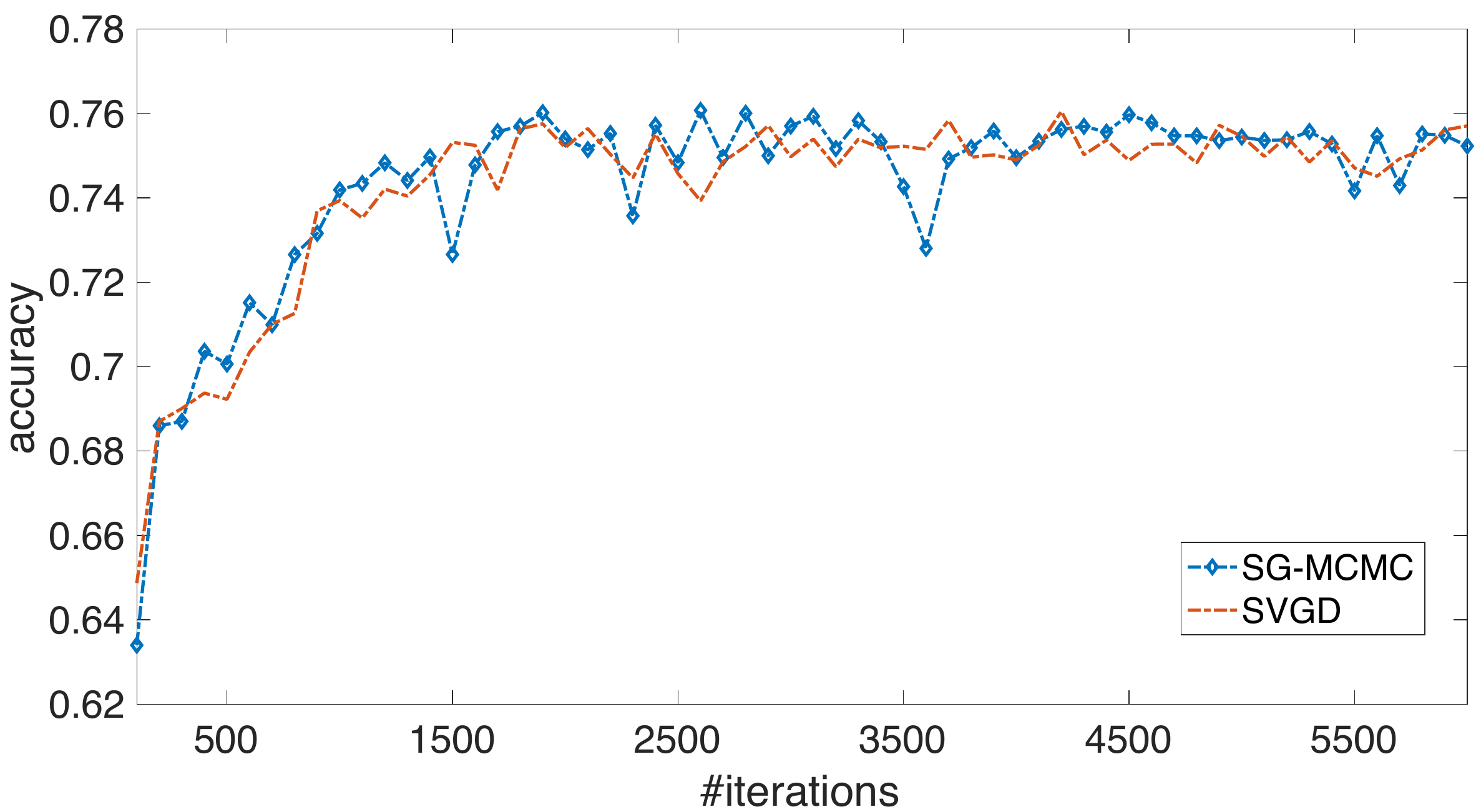}
	\hspace{0.1cm}
	\includegraphics[width=0.48\linewidth]{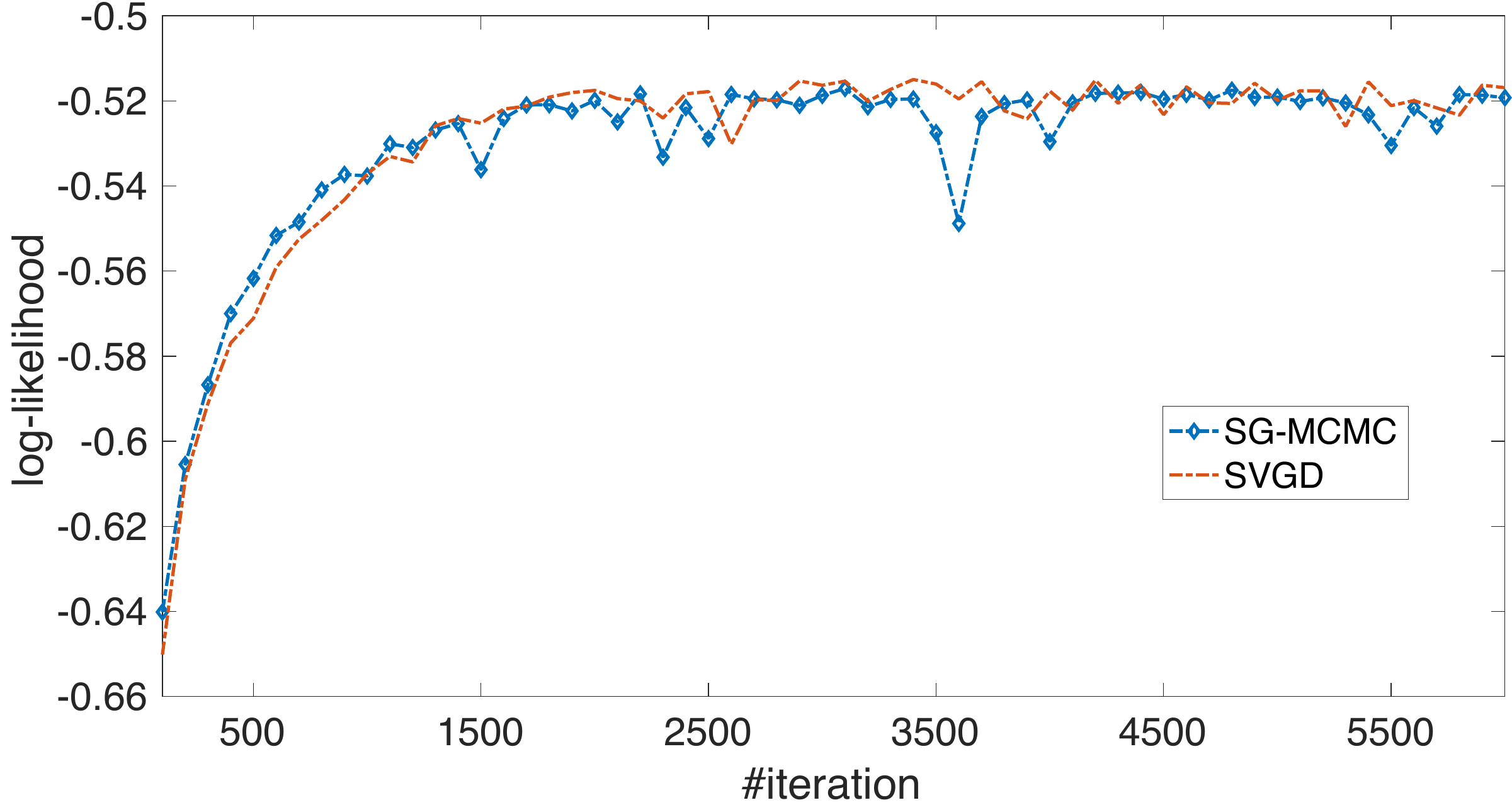}
	\caption{Test accuracies (left) and log-likelihoods (right) v.s.\! training iterations for SG-MCMC and SVGD.}\label{fig:lr_sgmcmc_svgd}
\end{figure}

We test our algorithm against SVGD on a simple logistic-regression task. We use the same model and data as \cite{LiuW:NIPS16}. For SVGD, we adopt the same setting as in \cite{LiuW:NIPS16}. Note we use the authors' implementation for SVGD \cite{LiuW:NIPS16}, which uses Adagrad for optimization, thus it is not strictly the standard SVGD (should perform better). For our particle optimization in SG-MCMC, we simply set the noise $\deltab$ to be standard isotropic normal. We scale $W^2_2\left(\tilde{\rho}_{k-1}, \rho\right)$ with 0.1, and use \eqref{eq:sgmcmc_grad} to update particles in SG-MCMC. Figure~\ref{fig:lr_sgmcmc_svgd} plots both test accuracies and test log-likelihoods w.r.t.\! the number of training iterations. It is clearly that without tuning, SG-MCMC already obtains slightly faster convergence speed than SVGD in terms of both test accuracy and test log-likelihood.

\section{Conclusion}
We propose novel methods to directly optimize particles in SG-MCMC, a paradigm to improve sample quality in standard SG-MCMC under limited samples. We develop our techniques by solving the Fokker-Planck equation of the correspond SG-MCMC on the space of probability distributions. A particular approximate solution of our framework results in SVGD with noisy gradient and momentum updates. To the best of our knowledge, this is the first time the relation of SG-MCMC and SVGD is established. A simple empirical result shows improvement of our method compared with standard SVGD. More extensive experiments are to be performed.


 \bibliographystyle{unsrt}
 \bibliography{reference}
 \clearpage
 \appendix
\end{document}